\documentclass{article}

\usepackage[nonatbib,final]{nips_2017}

\usepackage[utf8]{inputenc} 
\usepackage[T1]{fontenc}    
\usepackage{hyperref}       
\usepackage{url}            
\usepackage{booktabs}       
\usepackage{amsfonts}       
\usepackage{nicefrac}       
\usepackage{microtype}      

\usepackage{float} 
\usepackage{tikz}
\usepackage{tkz-euclide}
\usepackage{framed}
\usepackage{amssymb}
\usepackage{amsfonts}
\usepackage{amsmath}
\usepackage[amsmath,thmmarks]{ntheorem}
\usepackage{cite}

\usepackage{graphicx}
\usepackage{color}
\usepackage{pstricks}
\newrgbcolor{aasacolor}{.7 .1 .1}
\newrgbcolor{runecolor}{0.5, 0, 0.5}
\definecolor{eq-highlight}{rgb}{0, 0.5, 0}

\newboolean{DEBUG}
\setboolean{DEBUG}{true}

\ifthenelse {\boolean{DEBUG}}
{\newcommand{\aasa}[1]{{\bf \aasacolor{[Aasa says: #1]}}}}
{\newcommand{\aasa}[1]{}}

\ifthenelse {\boolean{DEBUG}}
{\newcommand{\rune}[1]{{\bf \runecolor{[Rune says: #1]}}}}
{\newcommand{\rune}[1]{}}

\theoremstyle{plain}
\theorembodyfont{\normalfont}

\newcounter{ex-counter}
\newtheorem{ex}[ex-counter]{Example}

\title{Learning from graphs with structural variation}

\author{
Rune Kok Nielsen$^*$, Andreas Nugaard Holm\thanks{Authors contributed equally to this work.}\ \ ,\ Aasa Feragen
\\ University of Copenhagen\\
\texttt{\{kok.nielsen,aholm,aasa\}@di.ku.dk}
}

\usepackage[ruled,linesnumbered]{algorithm2e}
\usepackage{amsmath}
\usepackage{graphicx}
\graphicspath{ {images/} }
\usepackage{listings}
\usepackage{rotating} 
\begin{document}

\maketitle
\begin{abstract}
We study the effect of structural variation in graph data on the predictive performance of graph kernels. To this end, we introduce a novel, noise-robust adaptation of the GraphHopper kernel and validate it on benchmark data, obtaining modestly improved predictive performance on a range of datasets. Next, we investigate the performance of the state-of-the-art Weisfeiler-Lehman graph kernel under increasing synthetic structural errors and find that the effect of introducing errors depends strongly on the dataset.
\end{abstract}
\section{Introduction}
Graph-structured data are abundant, e.g.~in social networks, bioinformatics, chemoinformatics and computer vision. In these settings, it is natural to expect structural variation even in samples with similar properties. For instance, graphs encoding image content may have additional or missing nodes due to occlusion, and social subnetworks of similar subjects will not be identical. However, classical graph mining algorithms quantify graphs as ''similar'' only if they contain many identical subgraphs, which is contrary to this intuition of natural variation. This leads to the question of whether structural variation in graphs reduces the predictive power of ''frequent subgraph'' features, and whether increased stability can be expected by encoding structural variation in the model.

Seeking to improve robustness in the face of structural variation, we introduce a \emph{gappy} version of the GraphHopper kernel~\cite{graphhopper}. The gappy GraphHopper kernel measures similarity between gappy~\cite{string_matching} shortest paths as opposed to regular shortest paths, see Ex.~\ref{ex:shortest_paths}. The gappy GraphHopper kernel can be efficiently computed, and modestly improves performance on a range of datasets. 
Our experiments indicate that the achievable improvement is limited as synthesizing structural errors in the graph does not severely degrade predictive performance -- meaning there is also not much to regain.
\begin{framed}
	\begin{minipage}[c]{0.7\linewidth}
		\begin{ex}
			\textbf{ Gappy Shortest Paths.}\ \ \ Consider the shortest paths starting in $a$ for the graph $G$ shown in $(i)$. There are four such paths, shown in $(ii)$: $\pi_1 = a;\ \pi_2=[a,b];\ \pi_3=[a,b,c]$ and $\pi_4=[a,b,c,d]$. By skipping one or more nodes in these paths we may add additional ''gappy'' paths to our pool of shortest paths rooted in $a$. For maximal gap size $s=1$ we may add the paths $\pi'_1=[a,c]$ and  $\pi'_2=[a,c,d]$ by skipping $b$ and $\pi_3'=[a,b,d]$ by skipping $c$, as shown in $(iii)$. For $s\geq 2$ we may also add $\pi_4'=[a,d]$ by skipping both $b$ and $c$.
			\vspace{2mm}
			
			This gives a richer set of features, which in this example would be robust to the case where either $b$ or $c$ is not present in otherwise similar graphs in the dataset. This is a situation that frequently occurs in graphs describing e.g.~image content or social interactions.
			
			\label{ex:shortest_paths}			
		\end{ex}
	\end{minipage}\begin{minipage}[c]{0.3\linewidth}
	\vspace{-0.45cm}
		\begin{figure}[H]
			\centering
			\includegraphics[width=0.7\linewidth]{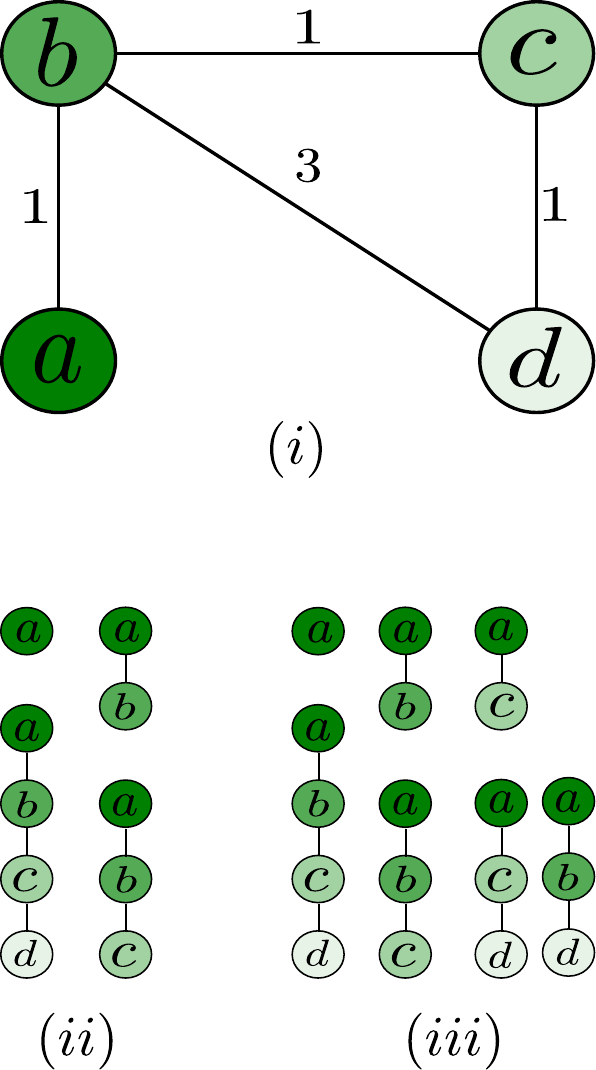}
		\end{figure}  
	\end{minipage}
\end{framed}

%
%

\section{A noise-robust graph kernel: The gappy GraphHopper kernel}
The GraphHopper kernel is an R-convolution kernel~\cite{Haussler99convolutionkernels} 
\[
k_{GH}(G,G') = \sum_{\pi \in P} \sum_{\pi' \in P'}k_p(\pi,\pi'),
\]
where $P$ and $P'$ are the families of shortest paths in the two graphs $G=(V,E)$ and $G'=(V',E')$, respectively. We assume graphs to be undirected with edge lengths $l(v,u)>0$ for $(v,u)\in E$. A path $\pi=(v_1,v_2,..,v_N)\in \Pi^G$ with $i^{th}$ node $\pi(i)$ and $|\pi|=N$ is a sequence of nodes such that the edge $(v_i,v_{i+1})\in E$ for $1\leq i < N$. The length of $\pi$ is defined as $l(\pi)=\sum_{i=1}^{N-1}l(v_i,v_{i+1})$ and a \textit{shortest} path $\pi=(v,..,u)$ is then a path of minimal length from $v$ to $u$. We denote the \textit{diameter} of $G$ by $\delta=\arg\max_{\pi\in P}|\pi|$.

In order to model structural variation we introduce the feature set $P^s$ containing gappy shortest paths, with each gap skipping at most $s\geq 0$ nodes (see Ex.~\ref{ex:shortest_paths}). We define \emph{the gappy GraphHopper kernel} $k_{GH}(G, G')$ as a sum of path kernels $k_p(\pi, \pi')$ over gappy shortest paths $\pi\in P^s,\pi'\in P'^s$:
\begin{equation}
\label{eq:k_graphhopper}
k_{GH}(G,G') = \sum_{\pi \in P^s} \sum_{\pi' \in P'^s}k_p(\pi,\pi'),
\end{equation}
where the path kernel $k_p(\pi, \pi')$ between paths with $|\pi| = |\pi'|$ is a sum of node kernels $k_n(v, v')$ over each pair $\pi(i), \pi'(i)$ for $1\leq i < N$, i.e.
\begin{align}
\label{eq:k_graphhopper_path}
k_p(\pi, \pi')=\begin{cases}
\sum_{i=1}^{|\pi|}k_n(\pi(i), \pi'(i)) & \text{for }|\pi|=|\pi'|,\\
0 & \text{otherwise.}
\end{cases}
\end{align}
Note that a gappy shortest path with $s=0$ is a regular shortest path such that $P^0=P$, such that~\eqref{eq:k_graphhopper} reduces to the original, gap-free GraphHopper kernel \cite{graphhopper} for $s=0$.

\begin{figure}[H]
	\centering
	\includegraphics[width=\linewidth]{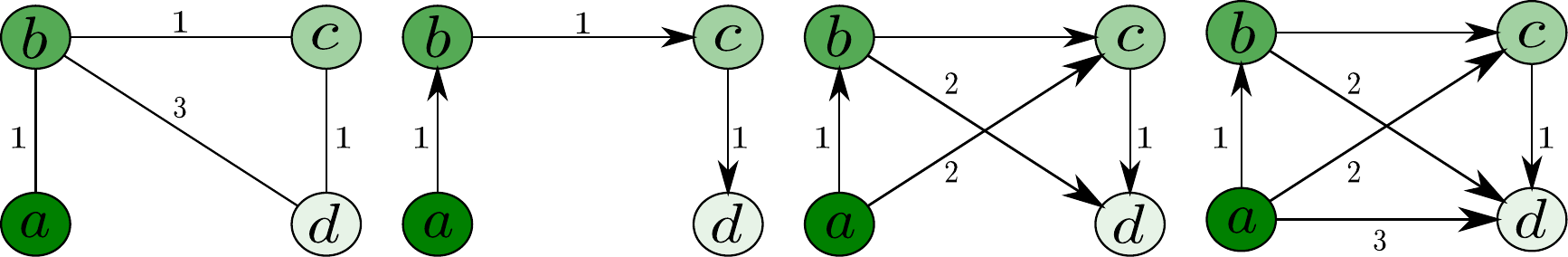}
	\caption{From the left: \textbf{First:} A simple graph $G$. \textbf{Second:} The shortest path DAG $G_a$ rooted in $a$. \textbf{Third:} $G_a^1$ with gaps of size $s\leq 1$. \textbf{Fourth:} $G_a^2$ with gaps of size $s\leq 2$.}
		\label{fig:shortest_path_dag_etc}
\end{figure}

An efficient algorithm for computing the gappy kernel follows from a series of observations, in a similar fashion as in the original kernel~\cite{graphhopper}. The \textbf{first} observation is that $k_{GH}$ can be written as a weighted sum of node kernels over pairs of vertices from the two graphs:
\begin{equation}
\label{eq:k_graphhopper_w}
k_{GH}(G,G') =  \sum_{v \in V} \sum_{v' \in V'} w(v,v')k_n(v,v').
\end{equation}
where $w(v,v')$ is the number of times $k_n(v,v')$ appears in Eq.~\ref{eq:k_graphhopper_path}.
The \textbf{second} observation is that $P^s = \bigcup_{a \in V} P_a^s$, 
where $P_a^s=\{\pi\in P^s | \pi(1)=a\}$. Using this property to decompose the kernel~\eqref{eq:k_graphhopper}, we obtain
\begin{align}
\begin{array}{ll}
k_{GH}(G,G') & = \sum_{\pi \in P^s} \sum_{\pi' \in P'^s}k_p(\pi,\pi')\\
& = \sum_{a \in V} \sum_{a' \in V'} \sum_{\pi \in P^s_a} \sum_{\pi' \in P'^s_{a'}}k_p(\pi,\pi')\\
& = \sum_{a \in V} \sum_{a' \in V'} \sum_{v \in V} \sum_{v' \in V'} w_{a, a'}(v, v') k_n(v, v')\\
& = \sum_{v \in V} \sum_{v' \in V'} \sum_{a \in V} \sum_{a' \in V'} w_{a, a'}(v, v') k_n(v, v'),
\end{array}
\end{align}
where $w_{a, a'}$ is the number of times $k_n(v,v')$ appears in \ref{eq:k_graphhopper_path} s.t.~$\pi(1)=a$ and $\pi'(1)=a'$. For $\pi(1)=a$ we say that $\pi$ \textit{is rooted }$a$ and define the subset of shortest paths rooted in $a$ as $P^s_a=\{\pi\in P^s | \pi(1)=a\}$. From this expression, we obtain
\begin{align}
w(v, v') = \sum_{a \in V} \sum_{a' \in V'} w_{a, a'}(v, v'),
\end{align}
The \textbf{third} observation is that the paths $\pi\in P^s_a$ form a gappy shortest path directed acyclic graph (DAG), similarly to the DAGs defined in \cite{graphhopper}, which allows us to efficiently compute $w_{a,a'}(v,v')$. The DAG formed by $\pi\in P^s_a$ is denoted $G^s_a=(V_a^s, E_a^s)$ and said to be rooted in $a$ (Fig \ref{fig:shortest_path_dag_etc}).

The \textbf{fourth} observation is that $w_{a,a'}$ can be efficiently computed through message-passing on the shortest path DAGs $G^s_a,G'^s_{a'}$. Let $\mathfrak{d}_a^v(i)$ be the count of paths in $G_a^s$ from $a$ to $v$ with discrete length $i$ and let $\mathfrak{o}_a^v(i)$ be the count of paths in $G^s_a$ starting in $v$ of discrete length $i$. The count of $v$ appearing as the $i^{th}$ node in a shortest path $\pi\in P_a^s$ with $|\pi|=j$ is then $\mathfrak{o}_a^v(i)\cdot\mathfrak{d}_a^v(j-i+1)$ and we may decompose $w_{a,a'}(v,v')$ as follows
\begin{align}
\begin{array}{ll}
w_{a, a'}(v, v') & = \sum_{i=1}^\delta \sharp \left\{ (\pi, \pi') \in P^s_a \times P'^s_{a'} \textrm{s.t.~} |\pi| = |\pi'| \textrm{ and } \pi(i) = v \textrm{ and }\pi'(i)=v' \right\}\\
& = \sum_{j=1}^\delta \sum_{i=1}^j \sharp \left\{ (\pi, \pi') \in P^s_a \times P'^s_{a'} \textrm{s.t.~} |\pi| = |\pi'| = j \textrm{ and } \pi(i) = v \textrm{ and }\pi'(i)=v' \right\}\\
& = \sum_{j=1}^\delta \sum_{i=1}^j \left( \mathfrak{o}^v_a(i) \cdot \mathfrak{d}^v_a(j-i+1) \right) ( \mathfrak{o}^{v'}_{a'}(i) \cdot \mathfrak{d}^{v'}_{a'}(j-i+1) )
\end{array}
\end{align}

As in~\cite{graphhopper}, the vectors $\mathfrak{d}_a^v$ and $\mathfrak{o}_a^v$ can be efficiently computed using message passing on $G_a^s$ in subquadratic time. Using a binary heap implementation of the Dijkstra algorithm for computing shortest path DAGs in \cite{graphhopper} and assuming connected graphs, the complexity of comparing $N$ graphs with the gap-free  GraphHopper kernel becomes $\mathcal{O}\left(N^2(n^2(d + \delta^2))+ N\left(n(\color{eq-highlight}m\color{black}\log n + m\delta) + n^2\delta^2\right)\right)$, where $n=|V|,m=|E|$, $\delta$ is the graph diameter and $d$ is the node feature dimension. The updated term is shown in green.

The gappy DAG $G_a^s$ is extended from the regular DAG $G_a^0$ before computing $\mathfrak{d}_a^v$ and $\mathfrak{o}_a^v$. In doing so, we traverse at most $s$ levels up through the ancestors of each node in the DAG; bypassing up to $s$ descendants. The amount of parents of a node is naturally limited by its degree (the number of edges adjacent to the node) -- as such, it is limited by the maximum degree $\Delta$ of the graph. A single node may then add at most $\Delta^{s+1}$ new edges to the DAG and adding gaps for each node in each DAG for one graph is then $\mathcal{O}(n^2\Delta^{s+1})$. Adding edges to the shortest path DAG $G_a$ will affect the complexity of computing $\mathfrak{o}_a^v$ and $\mathfrak{d}_a^v$ as the number of edges are no longer bound by $m$ but $m+n\Delta^{s+1}$. For $N$ graphs, this gives the updated complexity
\begin{align}
\mathcal{O}\left(N^2(n^2(d + \delta^2))+ N\left(n(m\log n + \left(m+\color{eq-highlight}{n\Delta^{s+1}}\color{black}{} \right)\delta) + n^2\delta^2 + \color{eq-highlight}{n^2\Delta^{s+1}}\color{black}\right)\right).
\end{align}
The exponential worst-case complexity of the new term seems daunting, but in practice $s$ should be relatively low and chosen from a short range of values as to limit growth of the parameter space.

\section{Experiments, results and discussion}

{\bf Experimental setup.} Classification experiments were carried out on the following datasets: ENZYMES\_SYMM~\cite{doi:10.1093/bioinformatics/bti1007,Schomburg_2017}, COLLAB~\cite{deep_graph_kernels}, MSRC$_9$~\cite{Neumann2016}, AIDS, COIL-RAG~\cite{Riesen2008,Nene96objectimage}, IMDB$_\text{binary}$\cite{deep_graph_kernels}, Letter-high~\cite{Riesen2008}, Letter-low~\cite{Riesen2008}, Letter-med~\cite{Riesen2008}, Mutagenicity~\cite{Riesen2008}, PROTEINS\cite{doi:10.1093/bioinformatics/bti1007,DOBSON2003771}, MUTAG~\cite{debnath}, MSRC\_21~\cite{Neumann2016} and MSRC\_21C~\cite{Neumann2016}. See Table~\ref{tab:prop} for their properties. Those graph datasets that did not have discrete node labels, were endowed with node degree as a discrete node label. 

Experiments were performed with the GraphHopper (GH) kernel without and with gaps, as well as with the Weisfeiler-Lehman (WL) kernel~\cite{weisfeiler_lehman}, which is the state-of-the-art frequent subgraph kernel for discretely labeled and unlabeled graphs. The GH kernel allows free choice of node kernel; a Dirac kernel was used when only discrete node labels were available, and a product of the Dirac and Gaussian node kernels when continuous-valued node attributes were available. The Gaussian bandwidth parameter was heuristically set to $\lambda = \frac{1}{d}$, where $d$ is the dimension of the continuous node attributes. All kernel matrices were normalized using the standard formula $k_{normalized}(G, G') = k(G, G')/\sqrt{k(G, G) k(G', G')}$. Classification was done with LibSVM~\cite{libsvm} using nested $10$-fold cross validation. The SVM $c$-parameter and the $s$ parameter of the gappy GH kernel were set using grid search in the inner loop, selecting $c$-parameters from the interval $[10^{-9}, 10^{-7}, \ldots, 10^9]$ and $s$ from the interval $[0, 1, \ldots, 5]$. All classification experiments were repeated 10 times, reporting the average accuracy.

\begin{figure}[t!]
	\includegraphics[width=1\linewidth]{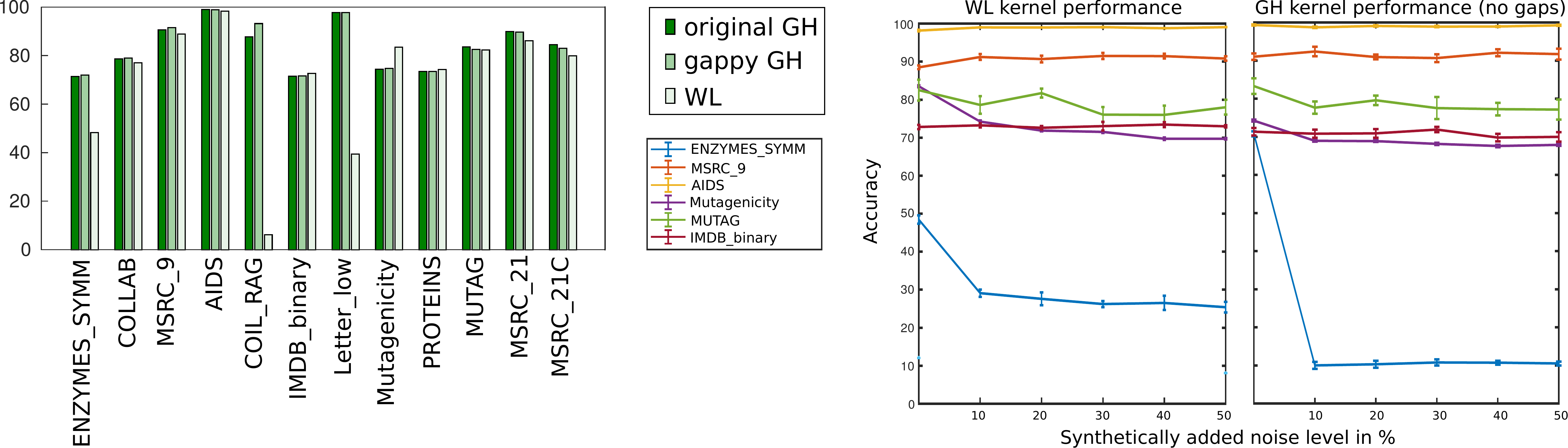}
	\caption{{\bf Left:} Mean classification accuracies for the GH kernel with and without gaps, as well as the WL kernel. {\bf Right:} Classification accuracy of WL kernel as a function of synthetically added noise in selected datasets.}
	\label{fig:plots}
\end{figure}

{\bf Results and discussion.} Fig.~\ref{fig:plots} (left) shows the mean accuracies of the three considered kernels in the benchmark datasets; note how the gappy GH is often a little better than the original GH, but note also that the effect is relatively small. This may in part be explained by our results in Fig.~\ref{fig:plots} (right), showing the mean classification accuracy of the WL and gap-free GH kernels with increasing synthetically added noise in the form of $x|V|$ randomly added nodes per graph, where $x\in [0,0.5]$. Note how some datasets are severely affected by the increased noise level, while others are not affected at all. Note also how the effect is the same with both the subgraph feature (WL) and the shortest-path feature (GH).

This difference in robustness may be caused by the ''triviality'' of some graph datasets~\cite{morteza}: If the discriminative signal is isolated in a few nodes, or if the node labels and graph structure are correlated, then the signal is likely to be robust in the presense of smaller structural alterations to the graphs.

\begin{table}[b]	


\centering
\caption{Properties of the used datasets.}
\resizebox{\textwidth}{!}{%
\begin{tabular}{|c|c|c|c|c|c|c|c|c|c|c|c|c|c|c|c}
\hline
Dataset & \rotatebox{90}{ENZYMES$_\text{symm}$} & \rotatebox{90}{MSRC\_9} & \rotatebox{90}{COIL-RAG} & \rotatebox{90}{COLLAB} & \rotatebox{90}{IMDB$_{\text{binary}}$} & \rotatebox{90}{PROTEINS} & \rotatebox{90}{AIDS}& \rotatebox{90}{Letter-high} & \rotatebox{90}{Letter-low} & \rotatebox{90}{Letter-med} & \rotatebox{90}{Mutagenicity} & \rotatebox{90}{MUTAG} & \rotatebox{90}{MSRC\_21} & \rotatebox{90}{MSRC\_21C}\\
\hline
Mean $|V|$ & 32.6 & 40.6 & 3.0 & 74.5 & 19.8 & 39.1 & 15.7 & 4.7 & 4.7 & 4.7 & 30.3 & 17.9 & 40.6 & 40.3\\
\hline
Mean $|E|$ & 62.1 & 97.9 & 3.0 & 2457.8 & 96.5 & 72.8 & 16.2 & 4.5 & 3.1 & 4.5 & 30.8 & 19.8 & 198.3 & 96.6\\
\hline
Density & 0.16 & 0.12 & 0.93 & 0.51 & 0.52 & 0.21 & 0.19 & 0.58 & 0.42 & 0.43 & 0.09 & 0.14 & 0.07 & 0.12\\
\hline
Discrete labels? & Yes & Yes & No & No & No & Yes & Yes & No & No & No & Yes & Yes & Yes & Yes\\
\hline
Vector attributes?  & 18 & No & 64 & No & No & 1 & No & 2 & 2 & 2 & No & No & No & No\\
Dimension? & & & & & & & & & & & & & & \\
\hline

\end{tabular}
}
 \label{tab:prop}
\end{table}

Note that the experimental setup above is not optimal for many of the datasets, as different choices of node kernels often give better classification accuracies. Note also that the WL kernel only utilizes discrete labels, and therefore in some cases uses less information than the GH kernels. However, as we are interested in the effect of modelling structural errors, we chose to only cross validate gap size and keep all other parameters fixed, as this was the only approach that scaled to the largest dataset COLLAB.

We note that while our investigation is performed in the context of graph kernels, the types of discriminative features used are generally the same across learning algorithms, and we thus expect our findings to generalize e.g.~to neural networks on graphs.

\section{Conclusion}

We have investigated the effect of structural variation on graph mining performance by two means: First, we introduced a \emph{gappy} shortest path feature which allows shortest paths to skip nodes along the way, and used these to develop a scalable gappy GraphHopper (GH) kernel. We compared the performance of the gappy GH to the original GH, and find improvement on several datasets, although mostly on a small scale. Next, to investigate why the improvement is small, we record the performance of the state-of-the-art WL kernel as a function of increasing synthetic noise. Here, we find different effects on different datasets -- on some, performance drops as one might expect, but on others, performance is retained or, in some cases, even get better. This difference in behavior may be related to the complexity of the graph datasets or the correlation between graph structure and node labels/attributes, and would be an interesting topic of discussion at the workshop.

\subsubsection*{Acknowledgements}
This research was supported by Lundbeck Foundation.
\small

\bibliography{references}{}
\bibliographystyle{plain}

\end{document}